\documentclass[10pt, a4paper]{article}
\usepackage{lrec2022} % this is the new LREC2022 Style
\usepackage{multibib}
\newcites{languageresource}{Language Resources}
\usepackage{graphicx}
\usepackage{tabularx}
\usepackage{soul}
\usepackage{titlesec}
\titleformat{\section}{\normalfont\large\bfseries\center}{\thesection.}{1em}{}
\titleformat{\subsection}{\normalfont\SmallTitleFont\bfseries\raggedright}{\thesubsection.}{1em}{}
\titleformat{\subsubsection}{\normalfont\normalsize\bfseries\raggedright}{\thesubsubsection.}{1em}{}
\renewcommand\thesection{\arabic{section}}
\renewcommand\thesubsection{\thesection.\arabic{subsection}}
\renewcommand\thesubsubsection{\thesubsection.\arabic{subsubsection}}
\usepackage{epstopdf}
\usepackage[utf8]{inputenc}

\usepackage{hyperref}
\usepackage{xstring}

\usepackage{color}

\usepackage{CJKutf8}
\usepackage{booktabs}
\usepackage[caption=false]{subfig}

\title{VISA: An Ambiguous Subtitles Dataset\\for Visual Scene-Aware Machine Translation}

\name{Yihang Li, Shuichiro Shimizu, Weiqi Gu, Chenhui Chu, Sadao Kurohashi} 

\address{Kyoto University, Kyoto, Japan \\
         \{liyh, sshimizu, gu, chu, kuro\}@nlp.ist.i.kyoto-u.ac.jp\\}

\abstract{
Existing multimodal machine translation (MMT) datasets consist of images and video captions or general subtitles, which rarely contain linguistic
ambiguity, making visual information not so effective to generate appropriate translations.
We introduce VISA, a new dataset that consists of $40$k Japanese--English parallel sentence pairs and corresponding video clips with the following key features: (1) the parallel sentences are subtitles from movies and TV episodes; (2) the source subtitles are ambiguous, which means they have multiple possible translations with different meanings; (3) we divide the dataset into \textit{Polysemy} and \textit{Omission} according to the cause of ambiguity. We show that VISA is challenging for the latest MMT system, and we hope that the dataset can facilitate MMT research. The VISA dataset is available at: https://github.com/ku-nlp/VISA.
\\ \newline }

\begin{document}
\begin{CJK}{UTF8}{min}
\maketitleabstract

\section{Introduction}

% talk about polysemy and omission

Neural machine translation (NMT) models relying on text data have achieved state-of-the-art performance for domains with little ambiguity in data. However, in several other domains, particularly real-time domains such as spoken language or sports commentary, the word sense ambiguity of verbs and nouns has a significant impact on translation quality \cite{gu-etal-2021-video}. Multimodal machine translation (MMT) focuses on using visual data as auxiliary information to tackle the ambiguity problem. The contextual information in the visual data assists in reducing the ambiguity of nouns or verbs in the source text data.

Previous MMT studies mainly focus on the image-guided machine translation (IMT) task \cite{elliott-etal-2016-multi30k,zhao-etal-2020-double}, where, given an image and a source sentence, the goal is to enhance the quality of translation by leveraging their semantic correspondence to the image. Resolving ambiguities through visual cues is one of the main motivations behind this task. However, videos are better information sources than images because one video contains an ordered sequence of frames and provides much more visual features such as motion features. 
Recently, some studies have started to focus on the video-guided machine translation (VMT) task \cite{wang2019vatex,hirasawa2020keyframe}.

VMT datasets face the problem of data scarcity. How2 \cite{sanabria2018how2} and VATEX \cite{wang2019vatex} datasets are recent efforts to alleviate the problem. In addition, previous datasets are limited to general subtitles or video captions which are actually descriptions of video clips. Besides the lack of practical use of caption MT, it has been shown that general subtitles and caption MT essentially do not require visual information due to the lack of language ambiguity in captions \cite{caglayan-etal-2019-probing}.

% The study of VMT is largely limited by dataset. However, there is a relative shortage of available datasets. How2 \cite{sanabria2018how2} and VATEX \cite{wang2019vatex} datasets are recent efforts to alleviate this bottleneck. In addition, previous datasets are limited to general subtitles or video captions which are actually descriptions of video clips. Besides the lack of practical use of caption MT, it has been shown that general subtitles and caption MT essentially does not require visual information due to the lack of language ambiguity in captions \cite{caglayan-etal-2019-probing}.

\begin{figure*}[!h]

\subfloat[\textit{Polysemy} example. ``放せ'' (let...go) can be regarded as a polysemy because it can be translated in two possible ways.]{%
  \includegraphics[clip,width=\textwidth]{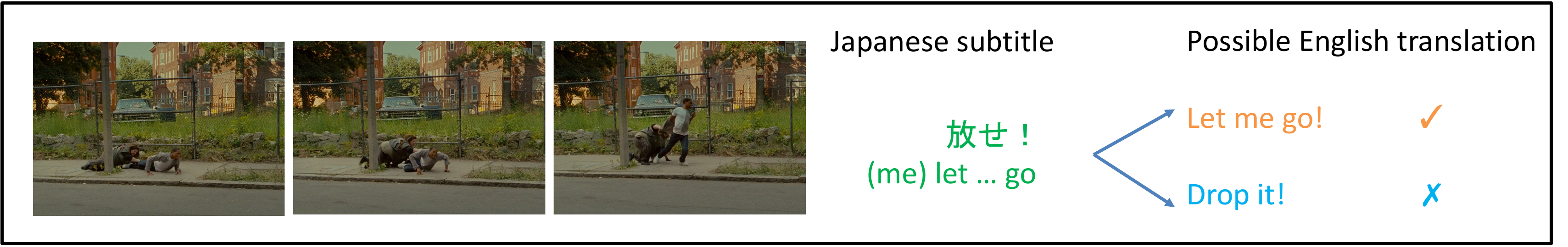}%
}

\subfloat[\textit{Omission} example. The subject in the Japanese subtitle is omitted.]{%
  \includegraphics[clip,width=\textwidth]{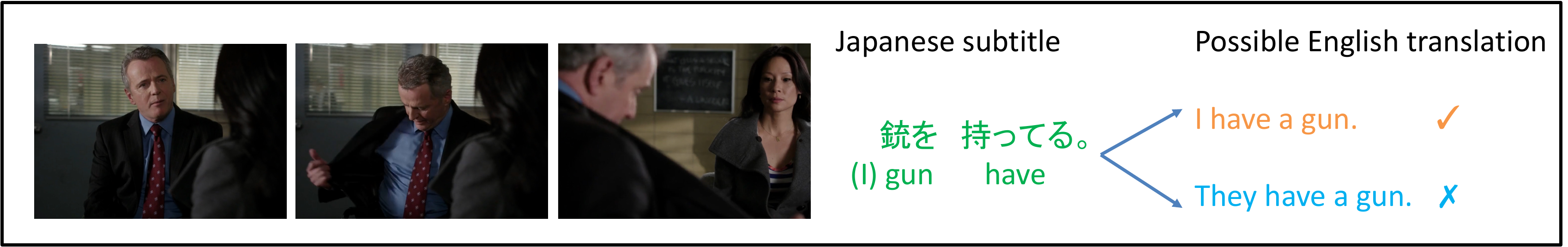}%
}

\caption{Examples of \textit{Polysemy} and \textit{Omission} data. In both cases, the translation can be disambiguated using the visual clue.}
\label{example}
\end{figure*}

To this end, we construct a new large-scale parallel subtitles dataset\footnote{The dataset is for research use only and is not available to the public. If you need the dataset, please send an email to the author.} for VMT research, VISA\footnote{VISA stands for VIsual Scene-Aware.}, which contains $39,880$ Japanese--English parallel subtitles pairs and corresponding video clips. 
Compared to previous large-scale VMT datasets, VISA is characterized by the following significant properties. (1) VISA uses subtitles from movies and TV series to construct parallel sentences for the translation task. Most of the subtitles are dialogues with short sentence lengths. (2) The source subtitles are all ambiguous subtitles, which means they have multiple possible translations with different meanings. In this case, it's difficult to determine which translation is correct without other information such as visual clues. (3) We focus on two causes of ambiguity, polysemy and omission, and divide the dataset accordingly. The ambiguity of \textit{Polysemy} subtitles is mainly caused by one or more polysemy in the subtitles. In contrast, the ambiguity of \textit{Omission} subtitles is mainly caused by the omission of the subject or object of subtitles. The examples are shown in Figure \ref{example}. The Polysemy part contains $20,666$ parallel subtitle pairs, while the \textit{Omission} part contains $19,214$ parallel subtitle pairs.

We conduct experiments on VISA with one of the latest VMT architecture proposed in \newcite{gu-etal-2021-video} as our baseline. We tested the performance of four models on the dataset according to this architecture, using only the text feature, with the motion feature, with the spatial feature, and with all three features, respectively. The results show that VISA is challenging for the existing VMT models.

In summary, our contributions are mainly three-fold:
\begin{itemize}
    \item We construct a new large-scale ambiguous parallel subtitles dataset to promote VMT research. The dataset is divided into the \textit{Polysemy} part and the \textit{Omission} part according to the cause of ambiguity.
    \item We conduct experiments on the VISA dataset with the latest VMT architecture to set a baseline of the dataset.
     \item Based on the analysis of the dataset and experimental results, we discuss problems and future perspectives for VMT models.
 \end{itemize}

% add example picture to show polysemy and omission. 

\section{Related Work} % 2

\textbf{Multimodal Machine Translation.} 
MMT involves drawing information from multiple modalities, assuming that the additional modalities will contain useful alternative views of the input data \cite{sulubacak2020multimodal}.
Previous studies mainly focus on IMT using images as the visual modality to help machine translation \cite{specia-etal-2016-shared,elliott-etal-2017-findings,barrault-etal-2018-findings}.
The utility of the visual modality has recently been disputed under specific datasets, or task conditions \cite{elliott-2018-adversarial,caglayan-etal-2019-probing}. However, using images in captions translation is theoretically helpful for handling grammatical characteristics and resolving translational ambiguities in translating between dissimilar languages \cite{sulubacak2020multimodal}.
VMT is a multimodal machine translation task similar to IMT but focuses on video clips rather than images associated with the textual input.
Depending on the textual context, there may be variations in video-guided translation. 
The source text can be speech transcripts from the video, which are often split as standard subtitles, or a description of the visual scene or activity displayed in the movie, which is frequently made for visually impaired persons \cite{sulubacak2020multimodal}.

\textbf{VMT Datasets.}
The scarcity of datasets is one of the biggest roadblocks to the advancement of VMT. Recent efforts to compile freely accessible data for VMT, such as the How2 \cite{sanabria2018how2}, and VATEX \cite{wang2019vatex} datasets, have begun to alleviate this bottleneck. 

The How2 dataset is a collection of $79,114$ instructional YouTube video clips (about $2,000$ hours) with associated English subtitles and summaries spanning a variety of topics. The average length of video clips is 90s. For multimodal translation, a $300$-hour subset of How2 subtitles that covers $22$ different topics is available with crowdsourced Portuguese translations. 
The Video and TeXt (VATEX) dataset (Wang et al. 2019b) is a bilingual collection of video descriptions containing over $41,250$ videos and $825,000$ captions in English and Chinese. With low-approval samples removed, the released version of the dataset includes $206,345$ translation pairs in total. 
However, both datasets do not require visual information due to the lack of language ambiguity.
We follow in the footsteps of How2 and VATEX, further utilize ambiguous subtitles to construct the new dataset, VISA, containing $39,880$ Japanese--English parallel subtitles pairs and corresponding $10$ seconds video clips, in total.

\section{Dataset Construction} % 3 

The main goal of our work is to construct an MMT dataset combined with parallel subtitles and corresponding video clips, in which the source subtitles must be ambiguous.
Our work uses Japanese--English parallel subtitles and considers Japanese and English as the source language and target language of translation tasks, respectively.

\begin{figure}[!h]
\begin{center}
\includegraphics[width=\columnwidth]{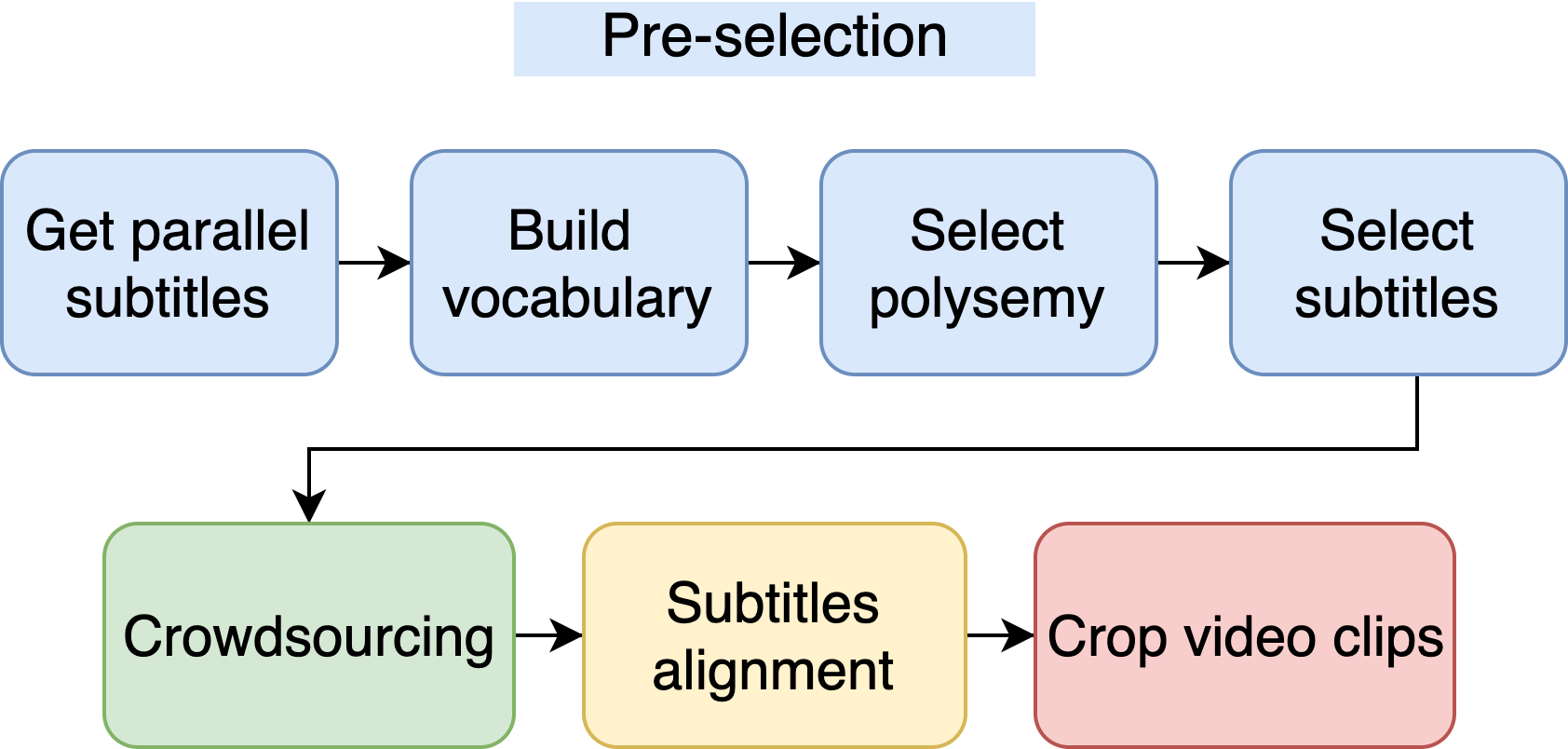} 

\caption{The procedure of dataset construction.}
\label{procedure}
\end{center}
\end{figure}

To construct the dataset, we select ambiguous parallel subtitles from the existing subtitle dataset and crop the corresponding video clips from movies or TV episodes. 
Subtitles selection is divided into two steps.
The first step is pre-selection. Because ambiguity is often caused by polysemy, we select Japanese subtitles that contain polysemy, which is more likely to be ambiguous. The second step is crowdsourcing. In this step, we use Yahoo Crowdsourcing\footnote{\url{https://crowdsourcing.yahoo.co.jp/}} to further select Japanese sentences that are indeed ambiguous. After selecting subtitles, we first align subtitles with videos, then crop video clips according to the timestamp of the aligned subtitle.
The whole procedure is shown in Figure \ref{procedure}.

\subsection{Pre-selection} % 3.1
Through pre-selection, we obtain subtitles containing polysemy in four steps: (1) get Japanese--English parallel subtitles; (2) build vocabulary from Japanese subtitles; (3) utilize BabelNet, a multilingual dictionary, to select polysemy from the vocabulary; (4) select subtitles which contain polysemy. 

\subsubsection{Get Parallel Subtitles}
For the first part, we get Japanese--English parallel subtitles from the OpenSubtitles \cite{lison-tiedemann-2016-opensubtitles2016} dataset. The main reason we chose Japanese as the source language is that it is a typical pro-drop language. OpenSubtitles is a subtitle dataset compiled from an extensive database of film and TV subtitles which includes a total of $1,689$ bitexts spanning $2.6$ billion sentences across $60$ languages. 
In OpenSubtitles, all parallel subtitles belonging to the same video are organized into a single subtitle file.
In addition to subtitles, OpenSubtitles also contains the Internet Movie Database (IMDb) id of the video corresponding to each subtitle file and the timestamp of each subtitle in the video.
From the OpenSubtitles dataset, we get the Japanese--English parallel subtitles, timestamps for each of them, and the corresponding video sources. In this way, we collected $1,213,468$ parallel subtitles pairs.

\subsubsection{Build Vocabulary}
After getting parallel subtitles, we can build a vocabulary for Japanese subtitles.
Since different forms of the same word have the same polysemous property, we reduce words to their original form before recording them into the vocabulary. In addition, a word may have different polysemous properties under different parts of speech (POS). In order to determine the polysemous property of words in a subtitle more precisely, we register each word into vocabulary with its POS. 
We use Juman++ \cite{tolmachev-etal-2018-juman} to segment Japanese subtitles into words, convert words to the original form, and check the POS of words. If a word is a content word (i.e., POS is one of noun, adjective, verb, or adverb), then we combine the original form and POS of the word into an entry of the vocabulary. For example, for a noun ``銃'' (gun) in the subtitle, we register ``銃｜noun'' into the vocabulary.

\subsubsection{Select Polysemy}
\paragraph{BabelNet polysemy}
As the third step, we aim to select polysemy from the vocabulary. 
Polysemy is a word or phrase which can represent different but related concepts.
Therefore, to determine whether a word is polysemous, we use BabelNet \cite{navigli-ponzetto-2010-babelnet} to check whether the word represents more than one concept.
BabelNet is an extensive, wide-coverage multilingual semantic network automatically constructed through a methodology that integrates lexicographic and encyclopedic knowledge from WordNet and Wikipedia.
In BabelNet, the resulting set of multilingual lexicalizations of a given concept is called a synset.
Therefore, if a word appears in more than one synset in BabelNet, we can regard it as a polysemy.

However, different synsets of BabelNet may represent very similar concepts. Therefore, we need to add restrictions to the process of determining polysemy. 
Due to the nature of BabelNet, a synset may contain lexicalizations from different sources. We can restrict the source to OmegaWiki and specify the POS of polysemy to reduce the number of synsets. 
In the sources provided by BabelNet, OmegaWiki has a clear division of words that represent different concepts, so it is more effective when used to determine polysemy.

\paragraph{Symilarity between synsets}
Furthermore, we compute the similarity between synsets and only preserve those words with small similarities. The algorithm we used to compute the similarity is Jiang and Conrath's similarity algorithm \cite{jiang-conrath-1997-semantic}, which uses WordNet as its central resource to quantify lexical semantic relatedness. It is found to be superior to other WordNet-based measures \cite{budanitsky-hirst-2006-evaluating}. We use BabelNet to export words' synsets in WordNet and then use this algorithm to calculate the synsets similarity. We consider a word as polysemous only if its minimum synsets similarity is less than a threshold of $0.06$, which we set based on experiments.

\paragraph{Removing specific words}
Specifically, we do not regard ``の'' (of) and ``来る'' (come) as polysemy. They have a high frequency in Japanese sentences but do not cause ambiguity in most situations. 

\subsubsection{Select Subtitles}
\paragraph{Select subtitles containing polysemy} At last, we select subtitles that contain polysemy. 
Again, we use Juman++ to segment the subtitles and determine the original form and the POS of each word in the sentence. Then we check whether the word-POS pair is in the polysemy vocabulary. We only keep subtitles that contain at least one polysemy.

\paragraph{Length ratio} Sometimes, the subtitles pairs are not parallel subtitles. For example, the source sentence is a complete Japanese sentence, while the target sentence is a phrase. In order to ensure that the subtitles pairs are parallel, we further select the subtitles with length ratio.
Given a source sentence $s$ and a target sentence $t$, we define a length ratio $r$ as:
\begin{equation}r(s, t)=\frac{\max(|s|, |t|)}{\min(|s|, |t|)}\end{equation}
in which $|x|$ is the number of words in sentence $x$. We only preserve subtitles where $r\leq2$.

\subsection{Crowdsourcing} % 3.2
After selecting subtitles containing polysemy, we further selected the subtitles that were indeed ambiguous on sentence level through crowdsourcing.
Because the purpose of crowdsourcing is to determine whether Japanese subtitles are ambiguous, for each group of duplicate Japanese subtitles, we only preserve one of them.

For each subtitle, we present it to workers together with annotations about the polysemy of the subtitle. 
Each annotation consists of polysemy and the main lemma of all synsets corresponding to the polysemy in both Japanese and English (e.g., 放す: 逃す/let\_go, 切り放す/let\_loose, 逃す/free, 見捨てる/abandon).

\begin{figure}[!h]
\begin{center}
\includegraphics[width=\columnwidth]{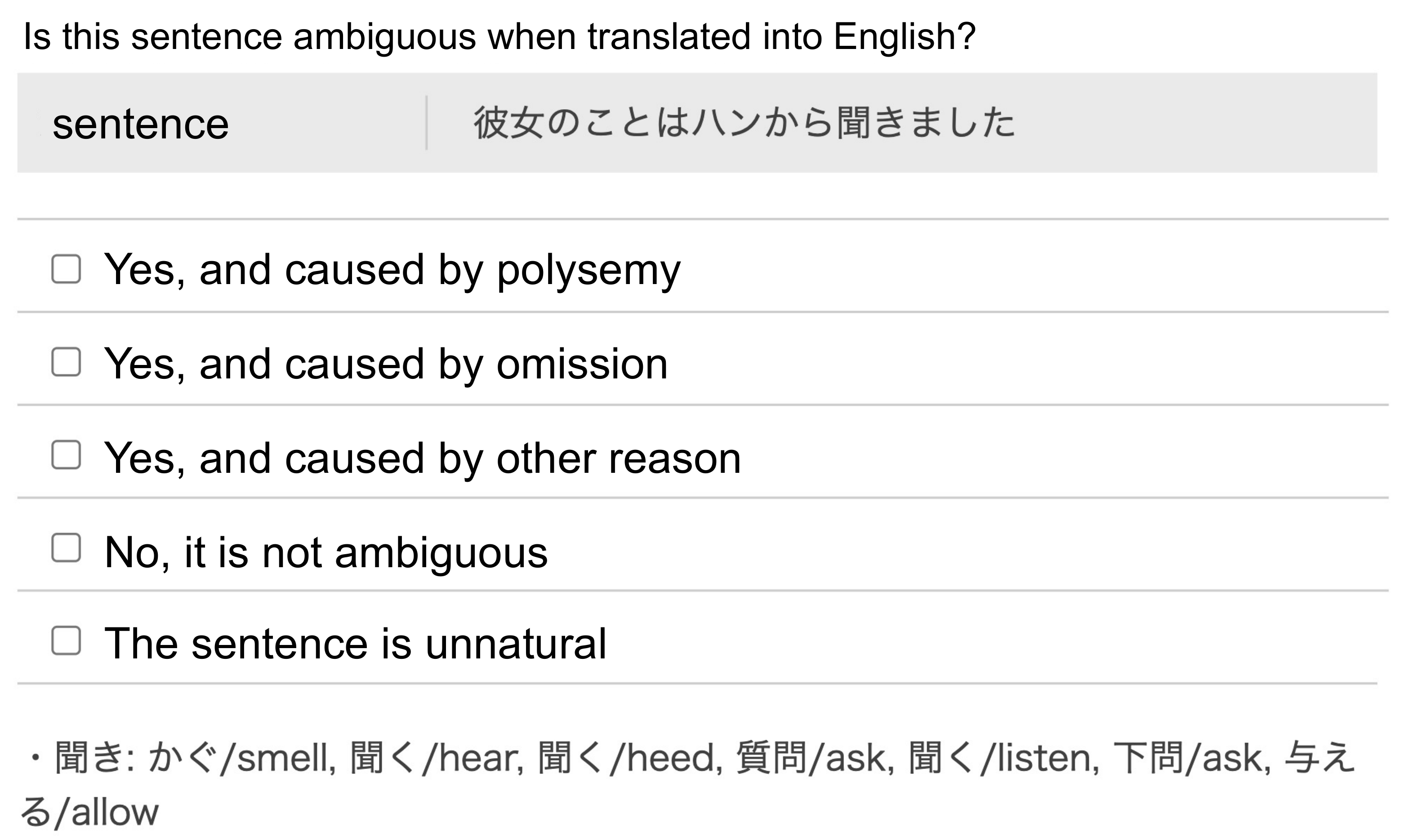} 

\caption{The interface for crowdsourcing.}
\label{interface}
\end{center}
\end{figure}

Furthermore, we ask workers whether the Japanese subtitle is ambiguous when translated into English and provide five choices to them: (A) yes, and caused by polysemy; (B) yes, and caused by omission; (C) yes, and caused by other reasons; (D) no, it is not ambiguous; (E) the sentence is unnatural.
Because sentence ambiguity may be caused by more than one reason, workers can choose multiple choices. The interface is shown in Figure \ref{interface}.
In these choices, we only use the ambiguous subtitles caused by polysemy and omission to construct the dataset and give up the ambiguous subtitles caused by other reasons. The main reason we provide this choice to workers is to ensure that the options cover all possibilities.

Three different workers answered each question. Each worker may choose multiple choices, and if an option is chosen by at least two workers of the three workers, we consider it a valid answer and consider the subtitle as a valid subtitle. 
Because the crowdsourcing was conducted on a large scale, we split it into multiple crowdsourcing tasks and posted one of them each day.

\begin{figure}[!h]
\begin{center}
\includegraphics[width=\columnwidth]{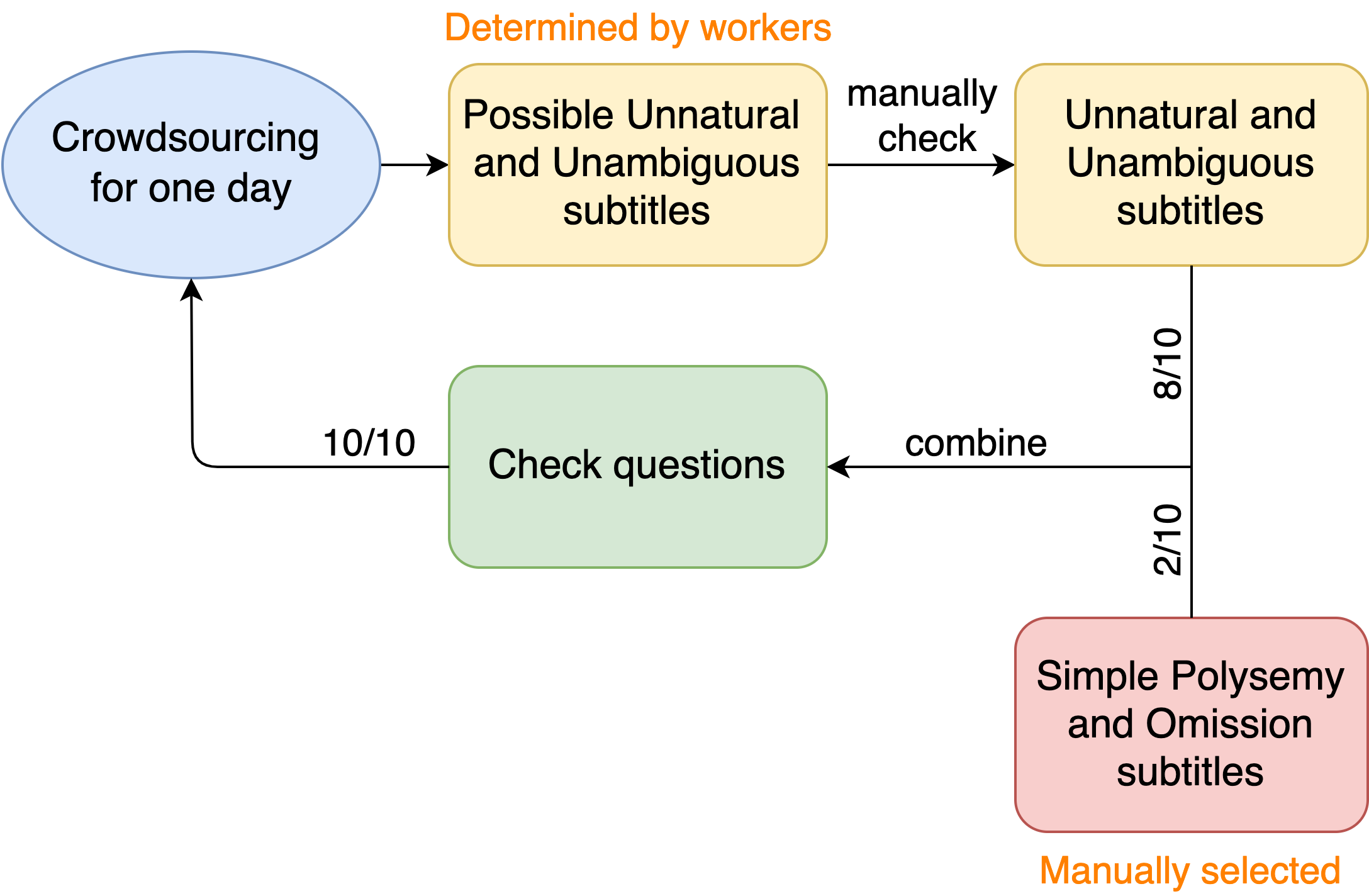} 

\caption{The procedure of check questions selection.}
\label{check question}
\end{center}
\end{figure}

To ensure the quality of workers' answers, we set up hidden check questions in crowdsourcing questions. We accepted a worker's answers only if s/he has correctly answered the check question. We select check questions using valid subtitles of completed tasks and manually selected questions as shown in Figure \ref{check question}. Because determining the ambiguity of sentences requires some associative ability, we mainly select unambiguous or unnatural sentences from the completed tasks to construct the check questions. However, to ensure the balance of the check questions, we also included two manually selected simple questions, a polysemy question and an omission question, in each day's check questions. By doing so, we ensured that these inspection questions were not too difficult but effective so that workers who answered carefully could be paid.

\subsection{Subtitles Alignment} % 3.3
% How to cite alass?
We collected $2,337$ videos corresponding to the subtitles. The videos are either movies or TV episodes. Different from video captions in which we do not need to care about timestamps, for subtitles, we must ensure that the timestamps of the subtitles correspond exactly to the time when the subtitles appear. However, many timestamps of the subtitles are not well aligned to the videos. For example, there may be a constant shift between the subtitles and the videos. In order to find the exact video clip corresponding to the subtitles, we need to align the original subtitles from the OpenSubtitles dataset with the videos.

We utilize alass\footnote{https://github.com/kaegi/alass} to align the original subtitles with videos. Alass can perform subtitles alignment in two ways. One is to align subtitle files with incorrect timestamps to subtitle files with correct timestamps, such as those that come with videos. The other is to align the incorrect subtitle file with the corresponding video using voice activity detection (VAD). The former performs significantly better and almost always aligns perfectly. After alignment, alass outputs a subtitle file with correct timestamps.

\begin{figure}[!h]
\begin{center}
%\fbox{\parbox{6cm}{
%This is a figure with a caption.}}
% old picture \includegraphics[scale=0.5]{lrec2020W-image1.eps} 
\includegraphics[width=\columnwidth]{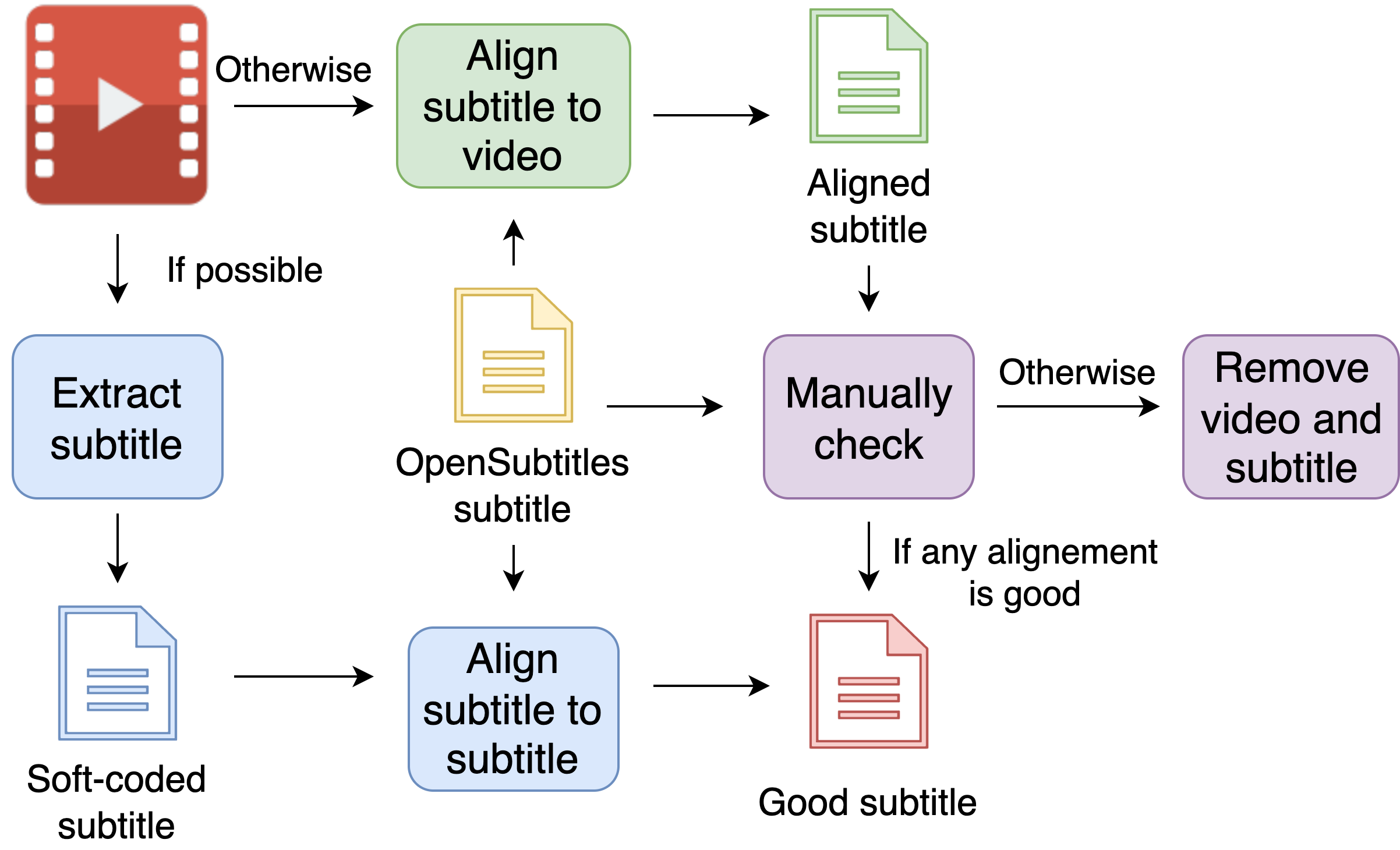} 

\caption{The procedure for subtitles alignment. The blue path is the first path, while the green and the purple path is the second path.}
\label{alignment}
\end{center}
\end{figure}

% add picture here
Based on the two functions of alass, we designed two subtitle alignment paths. The procedure is shown in Figure \ref{alignment}. The first one is based on aligning original subtitles to extracted subtitles with alass. Many videos contain soft-coded subtitles, which can be extracted from videos and are correct. Therefore, we utilize FFmpeg to extract soft-coded subtitles from videos and align original subtitles from the OpenSubtitles dataset with the extracted subtitles. As a result, we get well-aligned subtitles. 

If the video does not contain a soft-coded subtitle file, we need to perform the second path, which aligns the subtitles with the video. However, the aligned subtitles may still be incorrect. 
Therefore, we manually check if there is a well-aligned subtitle file in the aligned subtitle file and the original subtitle file. If both subtitle files are incorrect, we remove the video and the corresponding subtitles. 

In this way, we can align the subtitles and the videos to determine precisely where each subtitle appears in the video. This lays the groundwork for cropping the video clips corresponding to the subtitles.

\subsection{Video Cropping} % 3.4
With well-aligned subtitle files and videos, we can crop video clips for the ambiguous parallel subtitles. 
We standardized the format of the video clips as $10$-second $25$-fps mp4 files. 
From the midpoint of each subtitle's period, each video clip takes $5$ seconds before and after, respectively.
We only keep the video content of video clips and remove the audio content.
At last, we combine ambiguous parallel subtitles and corresponding video clips to construct the dataset.

\section{Dataset Statistics through Construction} % 4

\subsection{Pre-selection} % 4.1
The core task of the pre-selection is to build the polysemy vocabulary. As shown in Table \ref{pre-selection}, the number of subtitles corresponding to the polysemy vocabulary is reduced step by step during the selection process.
As a result, we select $4,875$ polysemy from the word-POS vocabulary of size $55,414$ and get $329,211$ parallel subtitles.

\begin{table}[!h]
\begin{center}
\begin{tabularx}{\columnwidth}{rr}

      \toprule
      \textbf{Select setting}&\textbf{Number of subtitles}\\
      \midrule
      Original & 1,213,468\\
      + BabelNet polysemy & 815,705\\
     + Synsets similarity & 352,344\\
     + Remove specific words& 329,211\\
     + Length ratio & 309,895\\
      \bottomrule

\end{tabularx}
\caption{The number of subtitles during pre-selection.}
\label{pre-selection}
 \end{center}
\end{table}

\subsection{Crowdsourcing} % 4.2
In crowdsourcing, we present five choices to workers.
We classify each subtitle according to how many (none, two, or three) of the three workers agree with its category. Because the questions are multiple-choice, a subtitle may belong to more than one category.
Only if a subtitle is not a valid subtitle, we classify it as no agreement. 
For example, if the three workers' choices are $\{A, B\}$, $\{A, B, C\}$, and $\{A\}$, the subtitle will belong to both $A$ of three agreements category and $B$ of two agreements category.
The statistics of the subtitles are shown in Table \ref{tab:agree}.
After keeping only those answers agreed by at least two workers, we get the valid subtitles as shown in Table \ref{tab:crowd-result}.
We can see that most of the subtitles are not ambiguous, while the number of polysemy part and omission part are similar.

After crowdsourcing, we only preserve the ambiguous subtitles caused by polysemy or omission and sample $50$ polysemy subtitles and $50$ omission subtitles from them. We asked two native Japanese speaker to help us manually check whether these subtitles are really ambiguous and what caused the ambiguity. As a result, $82\%$ \textit{Polysemy} subtitles and $88\%$ \textit{Omission} subtitles have correct categories after crowdsourcing.

% \begin{figure}[!h]
% \begin{center}
% \includegraphics[width=\columnwidth]{agree_of_sub.png} 

% \caption{The number of subtitles during crowdsourcing.}
% \label{agree}
% \end{center}
% \end{figure}

% no agreement: [28464 26843 8915 32787 14495], 
% 2 agreement: [22065, 20888, 966, 91769, 8851], 
% 3 agreement: [3099, 2818, 13, 40005, 1851]
% [25164, 23706, 979, 131774, 10702]

\begin{table}[h]
\centering
\begin{tabular}{@{}lrrr@{}}
\toprule
                & \multicolumn{3}{c}{\textbf{Agreement}} \\ \cmidrule(l){2-4} 
\textbf{Choice} & Three       & Two         & None       \\ \midrule
Polysemy        & 3,099       & 22,065      & 28,464     \\
Omission        & 2,818       & 20,888      & 26,843     \\
Other           & 13          & 966         & 8,915      \\
Not ambiguous   & 40,005      & 91,769      & 32,787     \\
Unnatural       & 1,851       & 8,851       & 14,495     \\ \bottomrule
\end{tabular}
\caption{The number of subtitles during crowdsourcing.}
\label{tab:agree}
\end{table}

\begin{table}[h]
    \centering
    \begin{tabular}{@{}lrr@{}}
    \toprule
    \textbf{Type}          & \textbf{Number}  & \textbf{Percentage} \\ \midrule
    Polysemy      & 25,164  & 13.08\%    \\
    Omission      & 23,706  & 12.33\%    \\
    Other         & 979     & 0.51\%     \\
    Not ambiguous & 131,774 & 68.52\%    \\
    Unnatural     & 10,702  & 5.56\%     \\ \bottomrule
    \end{tabular}
    \caption{The number of valid subtitles after crowdsourcing.}
    \label{tab:crowd-result}
\end{table}

\subsection{Subtitles Alignment} % 4.3
There are $2,751$ videos corresponding to the subtitles. We had $2,240$ videos of them at the beginning, and $1,978$ videos of them became well-aligned subtitles after subtitles alignment. The specific number of different alignments is shown in Table \ref{alignment number}. The result shows that most subtitles can be well aligned to videos, and subtitles alignment works better for movies rather than TV episodes. We only preserve well-aligned subtitles to construct the dataset.

\begin{table}[!h]
\begin{center}
\begin{tabularx}{\columnwidth}{lrrr}

      \toprule
      & \textbf{Movies} & \textbf{TV episodes} & \textbf{Total}\\
      \midrule
      Good alignment & 757 & 1,221 & 1,978\\
      Bad alignment & 57 & 205 & 262\\
      \midrule
      Good ratio & 93.00\% & 85.62\% & 88.30\%\\
      \bottomrule

\end{tabularx}
\caption{The number of videos during subtitles alignment.}
\label{alignment number}
 \end{center}
\end{table}

\subsection{Final Result} % 4.4
As a result, we constructed a dataset that consists of $39,880$ parallel subtitles and corresponding video clips.
This dataset is divided into two categories, \textit{Polysemy} and \textit{Omission}, based on what causes the subtitle ambiguity. The \textit{Polysemy} part has $20,666$ parallel subtitles, while the \textit{Omission} part has $19,214$ parallel subtitles.
There is a slight overlap between these two parts containing $1,971$ parallel subtitle pairs.
The final number of subtitles is more than the number of subtitles in the crowdsourcing result because we removed the duplicate Japanese subtitles before crowdsourcing, and we restored the duplicate subtitles after crowdsourcing.

\section{Experimental Settings} % 5
\subsection{Datasets} For the dataset, in addition to experimenting on the whole VISA dataset, we also conducted experiments on the \textit{Polysemy} and \textit{Omission} datasets separately. The splits of the three datasets are shown in Table \ref{tab:split}.

\begin{table}[]
\centering
\begin{tabular}{@{}lrrr@{}}
\toprule
\multicolumn{1}{c}{\textbf{Split}}    & \multicolumn{1}{c}{\textbf{Train}}  & \multicolumn{1}{c}{\textbf{Validation}} & \multicolumn{1}{c}{\textbf{Test}}  \\ \midrule
Polysemy & 18,666 & 1,000      & 1,000 \\
Omission & 17,214 & 1,000      & 1,000 \\
Combined  & 35,880 & 2,000      & 2,000 \\ \bottomrule
\end{tabular}
\caption{The splits of \textit{Polysemy}, \textit{Omission}, and combined dataset.}
\label{tab:split}
\end{table}

\subsection{Models} \label{sec:models}
To test the performance of MMT systems on our dataset, we implement the VMT translation (VMT) architecture described in \newcite{gu-etal-2021-video} and consider four models based on this architecture as our baseline models. 

The main idea of the VMT architecture is to use both spatial and motion representations in videos as auxiliary information to translate sentences. The architecture consists of the following four modules: (1) Text Encoder: A Bi-GRU \cite{650093} encoder transforms source sentences into text representations; (2) Motion Encoder: After extracting motion features with pretrained I3D \cite{DBLP:conf/cvpr/CarreiraZ17} model for action recognition, it uses a Bi-GRU motion encoder and a positional encoding (PE) layer to encode the motion features into ordered motion representations; (3) Spatial HAN: The architecture uses a hierarchical attention network to model the spatial information from object-level to video-level and extract contextual spatial representations; (4) Target Decoder: All representations from respective encoders are processed with attention mechanisms \cite{DBLP:conf/emnlp/LuongPM15} and help decode the target word embedding with a GRU decoding layer.

Based on this VMT architecture, we performed experiments with four models: (1) NMT model: only use the text encoder; (2) VMT (motion) model: use both the text and motion encoders; (3) VMT (spatial) model: use both the text encoder and spatial HAN; (4) VMT (both): use the whole architecture using text, motion, and spatial representations.

We trained the models on the three datasets following most of the model settings described in \newcite{gu-etal-2021-video}. As a special case, for the experiment on the \textit{Polysemy} dataset using VMT (both) model, we set drop out as $0.4$ instead of $0.5$ because it's difficult for the model to converge.

\subsection{Evaluation Metrics}
We adopt three automatic evaluation metrics: BLEU \cite{papineni-etal-2002-bleu}, METEOR \cite{banerjee-lavie-2005-meteor}, and RIBES \cite{isozaki-etal-2010-ribes}. 
The reason we introduced METEOR and RIBES scores as measures in addition to the BLEU score is that subtitles are similar to dialogue sentences and most subtitles are very short, with an average length of only around $7$ words. 

\begin{table*}[h]
\begin{center}
\begin{tabular}{c rrr rrr rrr}

      \toprule
      &  \multicolumn{3}{c}{\textbf{Polysemy}} & \multicolumn{3}{c}{\textbf{Omission}} & \multicolumn{3}{c}{\textbf{Combined}}\\
      \cmidrule(r){2-4} \cmidrule(r){5-7} \cmidrule(r){8-10} 
      \textbf{Model} & {\small BLEU} & {\small METEOR} & {\small RIBES} & {\small BLEU} & {\small METEOR} & {\small RIBES} & {\small BLEU} & {\small METEOR} & {\small RIBES} \\
      \midrule
      NMT & \textbf{7.71} & \textbf{24.67} & \textbf{14.37} & 5.58 & 18.54 & 9.86 & 13.14 & \textbf{29.00} & \textbf{20.66} \\
      VMT (motion) & 7.09 & 22.81 & 13.31 & 5.97 & 20.09 & \textbf{11.11} & \textbf{13.19} & 28.84 & 19.70 \\
      VMT (spatial) & 6.99 & 22.33 & 13.32 & \textbf{5.99} & 19.76 & 10.94 & 12.99 & 28.26 & 19.65 \\
      VMT & 7.40 & 22.33 & 12.64 & 5.63 & \textbf{20.10} & 10.98 & 12.89 & 28.45 & 19.45 \\
      \bottomrule

\end{tabular}
\caption{VMT results reported on BLEU, METEOR, and RIBES. Models are described in section \ref{sec:models}.}
\label{VMT}
\end{center}
\end{table*}

Because the calculation of the BLEU score is based on n-grams, the BLEU score cannot accurately reflect the quality of translation when the sentence length is very short. On the other hand, the METEOR score is based on the harmonic mean of unigram precision and recall, and the RIBES score is based on rank correlation coefficients modified with precision and mainly applied to distant language pairs such as Japanese and English. In practice, we utilize the Smooth 4 method described in \newcite{chen-cherry-2014-systematic} to correct the BLEU score.

% \begin{table*}[ht]
% \begin{center}
% \begin{tabular}{llrrrr}

%       \hline
%       \textbf{Score} & \textbf{Dataset} & \textbf{NMT} & \textbf{VMT (motion)} & \textbf{VMT (spatial)} & \textbf{VMT (motion + spatial)}\\
%       \hline
%           & Polysemy & \textbf{7.71}  & 7.09  & 6.99  & 7.40\\
%       BLEU & Omission & 5.58  & 5.97  & \textbf{5.99}  & 5.63\\
%           & Both      & 13.14 & \textbf{13.19} & 12.99 & 12.89\\
%       \hline
%           & Polysemy & \textbf{24.67} & 22.81 & 22.33 & 22.33\\
%       METEOR & Omission & 18.54 & 20.09 & 19.76 & \textbf{20.10}\\
%           & Both      & \textbf{29.00} & 28.84 & 28.26 & 28.45\\
%       \hline
%           & Polysemy & \textbf{14.37} & 13.31 & 13.32 & 12.64\\
%       RIBES & Omission & 9.86 & \textbf{11.11} & 10.94 & 10.98\\
%           & Both      & \textbf{20.66} & 19.70 & 19.65 & 19.45\\
%       \hline

% \end{tabular}
% \caption{Video-guided Machine Translation. Results are reported on the three evaluation metrics. Both: using the whole dataset consisting of both Polysemy and Omission datasets.}
% \label{VMT}
% \end{center}
% \end{table*}

\section{Results and Discussion} % 6
Table \ref{VMT} shows the results of the four baseline models evaluated using the three metrics. 
% Comparing the scores on different datasets, the scores on the whole VISA dataset are much higher than the scores on the Polysemy and Omission datasets. The size of the dataset has a great impact on translation performance. 
% Besides, which of the NMT model or VMT models performs better depends on the dataset.
Which of the NMT model or VMT models performs better depends on the dataset.
On the \textit{Omission} dataset, all three VMT models perform significantly better than the NMT model. The VMT model achieves $0.41$ BLEU score, $1.56$ METEOR score, and $1.25$ RIBES score over the NMT model, respectively. However, on the whole dataset and the \textit{Polysemy} dataset, the NMT model performs better than VMT models in most situations. Therefore, it is still challenging for the current VMT model to handle the polysemy problem. 

Why doesn't the current VMT model work well on VISA? By analyzing the translation results and the dataset, we found the following reasons. 

First, the videos do not necessarily contribute to the disambiguation. For example, many video clips only focus on the speaker's face when a subtitle appears. In movies and TV series, when a subtitle appears, the video tends to show the speaker's face. Although we can use this to get emotional information, it is more difficult to get other information such as object information from the video. Therefore, it is more challenging to translate the subtitles than to translate the descriptions of video clips.

The second reason is the lack of speaker recognition, which means we do not know who said the sentence. If we cannot determine who the speaker is, we cannot use the visual information of the speaker in the video. For example, in the omission example in Figure \ref{example}, if the speaker is the man with a gun, the omitted subject would be ``I'', while if the speaker is the woman, the omitted subject would be ``you''. In VISA, because the subtitles and the videos are well-aligned, we can try to determine the speaker based on the actions of the character's mouth.

Thirdly, in current VMT models, emotional information is not well captured. Object and spatial information are often not sufficient for disambiguation, and VMT models need to be able to obtain more information from the video. We observe that emotional information is necessary for disambiguation in many cases. For example, ``何言ってんだよ'' would be translated into "What are you talking about?" if the speaker is confused, while it would be translated into "It does not even make sense." if the speaker is angry or has no particular emotion. 

\section{Conclusion and Future Work}
In this paper, we introduced a new large-scale ambiguous parallel subtitles dataset for VMT research. The dataset is divided into two parts according to the cause of ambiguity. Experiments conducted on the dataset showed that the dataset is challenging for the latest VMT model. Based on the analysis of the dataset and experimental results, we pointed out possible directions to improve the VMT model.

However, there are still some problems with the dataset. The main problem is that although the subtitles are ambiguous, the videos do not necessarily contribute to the disambiguation. In the future, we plan to construct a test set for the VMT task and ensure that the videos in it can help disambiguate the ambiguous subtitles. And based on the test set, we plan to design a new VMT architecture that can obtain more information from video clips to help translate subtitles.

\section*{Acknowledgement}
This work was supported by ACT-I, JST.

% \nocite{*}
\section{Bibliographical References}\label{reference}
%\label{main:ref}

\bibliographystyle{lrec2022-bib}
\bibliography{lrec2022}

\end{CJK}
\end{document}